\documentclass[11pt]{article}
\usepackage{acl2016}
\usepackage{times}
\usepackage{latexsym}
\usepackage{amsmath}
\usepackage{amssymb}
\usepackage{amsthm} 
\usepackage{graphicx}
\usepackage{ifthen}
\usepackage[table]{xcolor}
\usepackage{tikz}
\usepackage{blkarray}
\usepackage{pgfplots}
\usepackage{rotating}
\usepackage{caption}
\usepackage{subcaption}
\usepackage{floatrow}
\usepackage{subcaption}
\usepackage{algorithm}
\usepackage{algpseudocode}
\usepackage{multirow}
\usepackage[hyphenbreaks]{breakurl}
\usepackage[hyphens]{url}

\aclfinalcopy

\usetikzlibrary{decorations.shapes,shapes.multipart,matrix,positioning,arrows,fit,calc}

\newcommand\sC{\ensuremath{\mathcal{C}}}

\newcommand\sF{\ensuremath{\mathcal{F}}}
\newcommand\sG{\ensuremath{\mathcal{G}}}

\newcommand\sM{\ensuremath{\mathcal{M}}}
\newcommand\sN{\ensuremath{\mathcal{N}}}

\DeclareMathOperator*{\tr}{tr}

\newcommand\R{\ensuremath{\mathbb{R}}}
\newcommand\Z{\ensuremath{\mathbb{Z}}}

\newcommand\eqdef{\ensuremath{\stackrel{\rm def}{=}}}

\newcommand{\bone}{\mathbf{1}}

\newcommand\refeqn[1]{(\ref{eqn:#1})}

\newcommand\refsec[1]{Section~\ref{sec:#1}}

\newcommand\reffig[1]{Figure~\ref{fig:#1}}

\newcommand\reftab[1]{Table~\ref{tab:#1}}

\newcommand\refalg[1]{Algorithm~\ref{alg:#1}}

\newcommand\refprop[1]{Proposition~\ref{prop:#1}}

\ifthenelse{\isundefined{\definition}}{\newtheorem{definition}{Definition}}{}
\ifthenelse{\isundefined{\assumption}}{\newtheorem{assumption}{Assumption}}{}
\ifthenelse{\isundefined{\hypothesis}}{\newtheorem{hypothesis}{Hypothesis}}{}
\ifthenelse{\isundefined{\proposition}}{\newtheorem{proposition}{Proposition}}{}
\ifthenelse{\isundefined{\theorem}}{\newtheorem{theorem}{Theorem}}{}
\ifthenelse{\isundefined{\lemma}}{\newtheorem{lemma}{Lemma}}{}
\ifthenelse{\isundefined{\corollary}}{\newtheorem{corollary}{Corollary}}{}
\ifthenelse{\isundefined{\alg}}{\newtheorem{alg}{Algorithm}}{}
\ifthenelse{\isundefined{\example}}{\newtheorem{example}{Example}}{}

\newcommand\citep\cite
\newcommand\citet\newcite

\newcommand\pointM{M_p}
\newcommand\ourM{\hat M}

\newcommand\fk[1]{\textcolor{blue}{[FK: #1]}}
\newcommand\nl[1]{\emph{#1}}
\newcommand\nlq[1]{``\emph{#1}''}
\newcommand\wl[1]{\texttt{#1}}

\newcommand{\sClp}{\sC_\text{\rm LP}}
\newcommand{\sCls}{\sC_\text{\rm LS}}
\newcommand{\sFlp}{\sF_\text{\rm LP}}
\newcommand{\sFls}{\sF_\text{\rm LS}}

\newcommand{\ns}{n_\text{s}}
\newcommand{\nt}{n_\text{t}}
\newcommand{\SF}{S}
\newcommand{\TF}{T}

\newcommand{\pZ}{\Z_{\scriptscriptstyle\geq0}}
\newcommand{\pR}{\R_{\scriptscriptstyle\geq0}}
\newcommand{\nmistakes}{n_\text{mistakes}}

\newcommand{\matindexw}[1]{\mbox{\begin{sideways}\scriptsize#1\end{sideways}}}

\tikzset{
    >=stealth',}

\title{Unanimous Prediction for 100\% Precision with Application to \\
Learning Semantic Mappings}

\author{
  Fereshte Khani \\
  Stanford University \\
  {\small\tt fereshte@cs.stanford.edu}
\And
  Martin Rinard \\
  MIT \\
  {\small\tt rinard@lcs.mit.edu}
\And
	Percy Liang \\
  Stanford University \\
  {\small\tt pliang@cs.stanford.edu}
}

\begin{document}

\maketitle

\begin{abstract}

Can we train a system that, on any new input, either says ``don't know'' or
makes a prediction that is guaranteed to be correct?
We answer the question in the affirmative provided our model family is well-specified.
Specifically, we introduce the unanimity principle:
only predict when all models consistent with the training data predict the same output.
We operationalize this principle for semantic parsing,
the task of mapping utterances to logical forms.
We develop a simple, efficient method that reasons over the infinite set of all consistent
models by only checking two of the models.
We prove that our method obtains 100\% precision even with a modest amount of
training data from a possibly adversarial distribution.
Empirically, we demonstrate the effectiveness of our approach on the standard GeoQuery dataset.

\end{abstract}

\section{Introduction}

If a user asks a system \nlq{How many painkillers should I take?},
it is better for the system to say ``don't know''
rather than making a costly incorrect prediction. 
When the system is learned from data, uncertainty pervades, and we must manage
this uncertainty properly  to achieve our precision requirement.
It is particularly challenging since training inputs might not be
representative of test inputs due to limited data,
covariate shift \citep{shimodaira2000improving},
or adversarial filtering \citep{nelson2009misleading,mei2015teaching}.
In this unforgiving setting,
can we still train a system that is \emph{guaranteed} to either abstain
or to make the correct prediction?

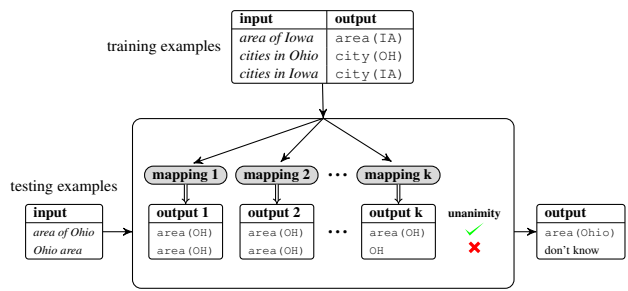
\begin{figure}
	\newcommand{\xscale}{0.5}	
	\newcommand{\yscale}{0.5}
	\newcommand{\scale}{0.5}
	\tikzset{
    >=stealth',
    inputBox/.style={
           rectangle,
			inner sep=0,
           rounded corners=0.1em,
           draw=black,
           text centered,
			scale=\scale},
   mappingBox/.style={
           rectangle,
			inner sep=0,
           rounded corners=0.3em,
           draw=black,
			fill=gray!30,
           text centered,
			scale=\scale},
    pil/.style={
           ->,
}
}

			\begin{tikzpicture}[xscale=\xscale,yscale=\yscale]

			\def\x{2.4}
			\def\y{3}
			\def\yy{1.5}

			\def\spx{0.2}
			\def\spy{0.2}

			\node (trainings)[inputBox] at (1.5*\x,\y+\yy+2*\spy){
				\begin{tabular}{l|l}
				\bf input & \bf output \\ 	\hline 
				\nl{area of Iowa}& \wl{area(IA)}\\
				\nl{cities in Ohio}&  \wl{city(OH)} \\
				\nl{cities in Iowa} & \wl{city(IA)}\\

				\end{tabular}
			};
			
			\node (A)[mappingBox] at (0,\yy){
				\begin{tabular}{l}
				\bf mapping 1 \\ 	
				\end{tabular}
			};

			\node (B) [mappingBox]at (\x,\yy){
				\begin{tabular}{l}
				\bf mapping 2 \\				
				\end{tabular}
			};
			
			\node (C) [mappingBox]at (2*\x+4*\spx,\yy){
				\begin{tabular}{l}
				 \bf mapping k \\ 
				\end{tabular}
			};

			\node (AA)[inputBox] at (0,0){
				\begin{tabular}{l}
				\bf output 1 \\ \hline
				\small\wl{area(OH)}\\
				\small\wl{area(OH)}\\

				\end{tabular}
			};
			
			\node (BB) [inputBox]at (\x,0){
				\begin{tabular}{l}
				\bf output 2 \\ \hline
				\small\wl{area(OH)}\\
				\small\wl{area(OH)}\\

				\end{tabular}
			};
			
			\node (CC) [inputBox]at (2*\x+4*\spx,0){
				\begin{tabular}{l}
				\bf output k \\ \hline
				\small\wl{area(OH)}\\
				\small\wl{OH}\\

				\end{tabular}
			};

			\node [left] at (trainings.west) {\tiny training examples};

			\node (testingsInput)[inputBox] at (-1*\x-3*\spx-\spx,0){
				\begin{tabular}{l}
				\bf input \\ \hline
				\small\nl{area of Ohio}\\ 
				\small\nl{Ohio area}\\ 

				\end{tabular}
			};
			
			\node (testingsOutput)[inputBox] at (4*\x+4*\spx,0){
				\begin{tabular}{l}
				\bf output \\ \hline
				\small\wl{area(Ohio)}\\ 
				\small don't know \\ 

				\end{tabular}
			};
			
			\node [above] at (testingsInput.north) {\tiny testing examples};

			\def\checkmark{\tikz\fill[fill=green,scale=0.5](0,.35) -- (.25,0) -- (1,.7) -- (.25,.15) -- cycle;} 
			
			\newcommand{\Cross}{$\mathbin{\tikz [line width=.5ex, red,scale=0.25] \draw (0,0) -- (1,1) (0,1) -- (1,0);}$}%
			
			\node (testingsResults)[xscale=\xscale,yscale=\yscale] at (3*\x + 2*\spx,0){
				\begin{tabular}{c}
				\small \bf unanimity \\ 
				\checkmark\\
				\Cross\\
				\end{tabular}
			};
			
			\node (3dots)[] at (1.5*\x+2*\spx,\yy){...};
			\node (3dots)[] at (1.5*\x+2*\spx,0){...};
			
    \node[scale=\scale,fit={(-\x/2-\spx,-\y/2) (3.6*\x,\y/2+\yy)}, inner sep=0pt, draw=black,rounded corners=.3em] (rect) {};

			\draw[pil, shorten >=3pt] (rect.north) -- (A.north);
			\draw[pil, shorten >=2pt] (rect.north) -- (B.north);
			\draw[pil, shorten >=3pt] (rect.north) -- (C.north);
			
			\draw[pil] (trainings.south) -- (rect.north);
			
			\draw[pil] (3.6*\x,0) -- (testingsOutput.west);
			\draw[pil] (testingsInput.east) -- (-\x/2-\spx, 0);

			\draw[double,-implies] (A.south) -- (AA.north);
			\draw[double,-implies] (B.south) -- (BB.north);
			\draw[double,-implies] (C.south) -- (CC.north);

\end{tikzpicture}
	\caption{ \label{fig:sound_parser}
    Given a set of training examples,
    we compute $\sC$, the set of all mappings consistent with the training examples.
    On an input $x$, if all mappings in $\sC$ unanimously predict the same output,
    we return that output; else we return ``don't know''.
	}
\end{figure}

Our present work is motivated by the goal of building reliable question answering systems and natural
language interfaces.  
Our goal is to learn a semantic mapping from examples of utterance-logical form pairs
(\reffig{sound_parser}).
More generally, we assume the input $x$ is a bag (multiset) of source atoms
(e.g., words $\{\nl{area},\nl{of},\nl{Ohio}\}$),
and the output $y$ is a bag of target atoms (e.g., predicates $\{\wl{area},\wl{OH}\}$).
We consider learning mappings $M$
that decompose according to the multiset sum:
$M(x) = \uplus_{s \in x} M(s)$ (e.g., $M(\{\nl{Ohio}\}) = \{\wl{OH}\}$, $M(\{\nl{area,of,Ohio}\}) = \{\wl{area,OH}\}$).
The main challenge is that an individual training example $(x,y)$ does not tell us which source
atoms map to which target atoms.\footnote{A semantic parser further
requires modeling the context dependence of words and the logical form structure joining
the predicates.
Our framework handles these cases with a different
choice of source and target atoms (see Section~\ref{sec:geo_query_experiments}).}

How can a system be 100\% sure about something if it has seen only a small number of possibly non-representative examples?
Our approach is based on what we call the \emph{unanimity} principle (Section~\ref{sec:unanimity_principle}).
Let $\sM$ be a model family that contains the true mapping from inputs to outputs.
Let $\sC$ be the subset of mappings that are consistent with the training data.
If all mappings $M \in \sC$ unanimously predict the same output on a test input, then we return that output;
else we return ``don't know'' (see Figure~\ref{fig:sound_parser}).
The unanimity principle provides robustness to the particular input distribution,
so that we can tolerate even adversaries \citep{mei2015teaching},
provided the training outputs are still mostly correct.

To operationalize the unanimity principle,
we need to be able to efficiently reason about the predictions of all consistent mappings $\sC$.
To this end, we represent a mapping as a matrix $M$,
where $M_{st}$ is number of times target atom $t$ (e.g., \wl{OH})
shows up for each occurrence of the source atom $s$ (e.g., \nl{Ohio}) in the input.
We show that unanimous prediction can be performed by
solving two integer linear programs.
With a linear programming relaxation (Section~\ref{sec:linear_algebraic_formulation}),
we further show that checking unanimity over $\sC$ can be done very efficiently
without any optimization but rather
by checking the predictions of just two random mappings,
while still guaranteeing 100\% precision with probability $1$ (Section~\ref{sec:linear_program}).

We further relax the linear program to a linear system, which gives us a geometric view of the unanimity:
We predict on a new input if it can be expressed as a ``linear combination'' of the training inputs.
As an example, suppose we are given training data consisting of
(CI) \nl{cities in Iowa},
(CO) \nl{cities in Ohio}, and
(AI) \nl{area of Iowa}
(Figure~\ref{fig:sound_parser}).
We can compute (AO) \nl{area of Ohio} by analogy: (AO) = (CO) - (CI) + (AI).
Other reasoning patterns fall out from more complex linear combinations.

We can handle noisy data (Section~\ref{sec:remove_noise}) by asking for
unanimity over additional slack variables.
We also show how the linear algebraic formulation enables other
extensions such as learning from denotations (\refsec{learning_from_denotations}),
active learning (\refsec{active_learning}),
and paraphrasing (\refsec{paraphrasing}).
We validate our methods in Section~\ref{sec:experiments}.
On artificial data generated from an adversarial distribution with noise,
we show that unanimous prediction obtains 100\% precision,
whereas point estimates fail.
On GeoQuery \citep{zelle96geoquery}, a standard semantic parsing dataset,
where our model assumptions are violated, we still obtain 100\% precision.
We were able to reach 70\% recall on recovering predicates and 59\% on full logical forms.

\section{Setup}
\label{sec:setup}
\begin{figure}
\def\scale{0.62}
\centering
\begin{tikzpicture}
\tikzset{
    myarrow/.style={->, >=latex', shorten >=2pt},
	mybox/.style={
	scale=\scale,
	rounded corners=.1em, 
	draw=black,
	inner sep=1,
	text width=3cm,
},
}
\def\x{2.1}
\def\y{2.5}
\def\btw{5}

\node (sourceAtoms and targetAtoms)[scale=\scale] at (1.5*\x,1.9*\y){
	\begin{tabular}{|l|l|}
	\hline \bf source atoms & \bf target atoms \\ \hline
	\{\nl{area}, \nl{of}, \nl{Iowa}\}& \{\wl{area}, \wl{IA}\}\\
	\{\nl{cities}, \nl{in}, \nl{Ohio}\}&  \{\wl{city}, \wl{OH}\} \\
	\{\nl{cities}, \nl{in}, \nl{Iowa}\} & \{\wl{city}, \wl{IA}\}\\
	\hline
	\end{tabular}
};

	\node (A)[mybox] at (0,\y){
		\begin{tabular}{p{0.5cm}p{0.1cm}p{1.15cm}}
	\multicolumn{3}{c}{\bf mapping 1} \\ \hline
		 \nl{cities}&$\rightarrow$&\{\wl{city}\} \\
	 \nl{in}&$\rightarrow$&\{\} \\
		 \nl{of}&$\rightarrow$  &\{\} \\
		 \nl{area}&$\rightarrow$ & \{\wl{area}\} \\
		 \nl{Iowa}&$\rightarrow$ & \{\wl{IA}\} \\
		\nl{Ohio}&$\rightarrow$ & \{\wl{OH}\} \\
		\end{tabular}
	};
	
	\node (B) [mybox]at (\x,\y){
			\begin{tabular}{p{0.5cm}p{0.1cm}p{1.15cm}}
\multicolumn{3}{c}{\bf mapping 2} \\ \hline
		 \nl{cities}&$\rightarrow$&\{\wl{}\} \\
	 \nl{in}&$\rightarrow$&\{\wl{city}\} \\
		 \nl{of}&$\rightarrow$  &\{\} \\
		 \nl{area}&$\rightarrow$ & \{\wl{area}\} \\
		 \nl{Iowa}&$\rightarrow$ & \{\wl{IA}\} \\
		\nl{Ohio}&$\rightarrow$ & \{\wl{OH}\} \\
		\end{tabular}
	};

	\node (C) [mybox]at (2*\x ,\y){
			\begin{tabular}{p{0.5cm}p{0.1cm}p{1.15cm}}
\multicolumn{3}{c}{\bf mapping 3} \\ \hline
		 \nl{cities}&$\rightarrow$&\{\wl{city}\} \\
	 \nl{in}&$\rightarrow$&\{\} \\
		 \nl{of}&$\rightarrow$  &\{\wl{area}\} \\
		 \nl{area}&$\rightarrow$ & \{\wl{}\} \\
		 \nl{Iowa}&$\rightarrow$ & \{\wl{IA}\} \\
		\nl{Ohio}&$\rightarrow$ & \{\wl{OH}\} \\
		\end{tabular}
	};

	\node (D) [mybox]at (3*\x ,\y){
		\begin{tabular}{p{0.5cm}p{0.1cm}p{1.15cm}}
\multicolumn{3}{c}{\bf mapping 4} \\ \hline
		 \nl{cities}&$\rightarrow$&\{\wl{}\} \\
	 \nl{in}&$\rightarrow$&\{\wl{city}\} \\
		 \nl{of}&$\rightarrow$  &\{\wl{area}\} \\
		 \nl{area}&$\rightarrow$ & \{\wl{}\} \\
		 \nl{Iowa}&$\rightarrow$ & \{\wl{IA}\} \\
		\nl{Ohio}&$\rightarrow$ & \{\wl{OH}\} \\
		\end{tabular}
	};

	\draw[myarrow, bend right] (sourceAtoms and targetAtoms.south) to [out=-8,in=-160] (A.north);
	\draw[myarrow] (sourceAtoms and targetAtoms.south) -- (B.north);
	\draw[myarrow] (sourceAtoms and targetAtoms.south) -- (C.north);
	\draw[myarrow] (sourceAtoms and targetAtoms.south) to [out=0,in=160] (D.north);

\end{tikzpicture}
\caption{\label{fig:atoms_and_mappings}
Given the training examples in the top table,
	  there are exactly four mappings consistent with these training examples.
}
\end{figure}
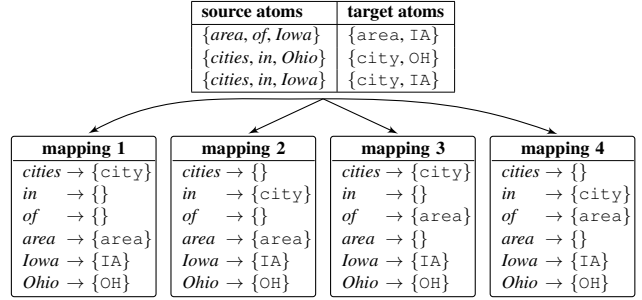
We represent an input $x$ (e.g., \nl{area of Ohio}) as a bag (multiset) of
\emph{source atoms} and an output $y$ (e.g., \wl{area(OH)}) as a bag of \emph{target atoms}.
In the simplest case, \emph{source atoms} are words and \emph{target atoms} are predicates---see
Figure~\ref{fig:atoms_and_mappings}(top) for an example.\footnote{Our semantic
  parsing experiments (Section~\ref{sec:geo_query_experiments}) use
more complex source and target atoms to capture some context and structure.}
We assume there is a true mapping $M^*$ from a source atom $s$ (e.g., \nl{Ohio})
to a bag of target atoms $t = M^*(s)$ (e.g., $\{\wl{OH}\}$).
Note that $M^*$ can also map a source atom $s$ to no target atoms ($M^*(\nl{of}) = \{\}$) or
multiple target atoms ($M^*(\nl{grandparent}) = \{\wl{parent}, \wl{parent}\}$). 
We extend $M^*$ to bag of source atoms via multiset sum: $M^*(x) = \uplus_{s \in x} M^*(s)$.

Of course, we do not know $M^*$ and must estimate it from training data.
Our training examples are input-output pairs
$D = \{(x_1, y_1), \dots, (x_n,y_n)\}$.
For now, we assume that there is no noise so that $y_i = M^*(x_i)$;
Section~\ref{sec:remove_noise} shows how to deal with noise.
Our goal is to output a \emph{mapping} $\ourM$ that maps
each input $x$ to either a bag of target atoms or ``don't know.''
We say that $\ourM$ has 100\% precision if 
$\ourM(x) = M^*(x)$ whenever $\ourM(x)$ is not ``don't know.''
The chief difficulty is that the source atoms $x_i$ and the target atoms $y_i$ are unaligned.
While we could try to infer the alignment,
we will show that it is unnecessary for obtaining 100\% precision.

\begin{figure*}
\newcommand{\matindex}[1]{\mbox{\scriptsize#1}}
\newcommand{\scc}{0.85}
	\begin{center}
		\scalebox{\scc}{
			$S=\begin{blockarray}{lcccccc}
			&\matindexw{\nl{area}} & \matindexw{\nl{of}} & \matindexw{\nl{Ohio}} & \matindexw{\nl{cities}} & \matindexw{\nl{in}} & \matindexw{\nl{Iowa}}\\	
			\begin{block}{l[cccccc]}
			\matindex{\nl{area of Iowa}} & 1 & 1 & 0 & 0 & 0 & 1 \\
			\matindex{\nl{cities in Ohio}} & 0 & 0 & 1 & 1 & 1 & 0 \\
			\matindex{\nl{cities in Iowa}} & 0 & 0 & 0 & 1 & 1 & 1 \\
			\end{block}
			\end{blockarray}$}
		\hspace{0.1cm}
		\scalebox{\scc}{
			$M=\begin{blockarray}{ccccc}
			&\matindexw{\wl{area}} & \matindexw{\wl{city}} & \matindexw{\wl{OH}} &  \matindexw{\wl{IA}}\\	
			\begin{block}{c[cccc]}
			\matindex{\nl{area}}		& 1 & 0 & 0 & 0\\
			\matindex{\nl{of}} 		& 0 & 0 & 0 & 0\\
			\matindex{\nl{Ohio}}		& 0 & 0 & 1 & 0 \\
			\matindex{\nl{cities}}	& 0 & 1 & 0 & 0 \\
			\matindex{\nl{in}}		& 0 & 0 & 0 & 0\\
			\matindex{\nl{Iowa}}		& 0 & 0 & 0 & 1\\
			\end{block}
			\end{blockarray}
			$}
		\hspace{0.1cm}
		\scalebox{\scc}{
			$T=\begin{blockarray}{lcccc}
			&\matindexw{\wl{area}} & \matindexw{\wl{city}} & \matindexw{\wl{OH}} & \matindexw{\wl{IA}}\\	
			\begin{block}{l[cccc]}
			\matindex{\wl{area(IA)}} & 1 & 0 & 0 & 1\\
			\matindex{\wl{city(OH)}} & 0 & 1 & 1 & 0\\
			\matindex{\wl{city(IA)}} & 0 & 1 & 0 & 1\\
			\end{block}
			\end{blockarray}$}
	\end{center}
	\caption{\label{fig:SandTandM}
    Our training data encodes a system of linear equations $S M = T$,
    where the rows of $\SF$ are inputs,
    the rows of $\TF$ are the corresponding outputs,
    and $M$ specifies the mapping between source and target atoms.
}
\end{figure*}

\subsection{Unanimity principle}
\label{sec:unanimity_principle}

Let $\sM$ be the set of mappings (which contains the true mapping $M^*$).
Let $\sC$ be the subset of mappings consistent with the training examples. 
\begin{align}
  \label{eqn:consistent}
  \sC &\eqdef \{ M \in \sM \mid M(x_i) = y_i, \forall i = 1, \dots, n \} 
\end{align}
Figure~\ref{fig:atoms_and_mappings} shows the four mappings consistent with the
training set in our running example.
Let $\sF$ be the set of \emph{safe inputs},
those on which all mappings in $\sC$ agree:
\begin{align}
  \label{eqn:safe}
  \sF &\eqdef \{ x : |\{ M(x) : M \in \sC \}| = 1 \}.
\end{align}
The \emph{unanimity principle}
defines a mapping $\hat{M}$ that returns the unanimous output on $\sF$
and ``don't know'' on its complement.  This choice obtains the following strong
guarantee:

\begin{proposition}
For each safe input $x \in \sF$, we have $\hat M(x) = M^*(x)$.
In other words, $M$ obtains 100\% precision.
\end{proposition}
Furthermore, $\hat M$ obtains the best possible recall given this model family subject
to 100\% precision, since for any $x \not\in \sF$ there are at least two possible outputs
generated by consistent mappings, so we cannot safely guess one of them.

\section{Linear algebraic formulation}
\label{sec:linear_algebraic_formulation}

To solve the learning problem laid out in the previous section,
let us recast the problem in linear algebraic terms.
Let $\ns$ ($\nt$) be the number of source (target) atom types.
First, we can represent the bag $x$ ($y$) as a $\ns$-dimensional ($\nt$-dimensional) row vector of counts;
for example,
the vector form of \nlq{area of Ohio} is $\begin{blockarray}{cccccc}
			\matindexw{\nl{area}} & \matindexw{\nl{of}} & \matindexw{\nl{Ohio}} & \matindexw{\nl{cities}} & \matindexw{\nl{in}} & \matindexw{\nl{Iowa}}\\	
			\begin{block}{[cccccc]}
			1 & 1 & 1 & 0 & 0 & 0 \\
			\end{block}
			\end{blockarray}$.
We represent the mapping $M$ as a non-negative integer-valued matrix,
where $M_{st}$ is the number of times target atom $t$ appears in the bag that source atom $s$ maps to (\reffig{SandTandM}).
We also encode the $n$ training examples as matrices:
$\SF$ is an $n \times \ns$ matrix where the $i$-th row is $x_i$;
$\TF$ as an $n \times \nt$ matrix where the $i$-th row is $y_i$.
Given these matrices, we can rewrite the set of consistent mappings \refeqn{safe} as:
\begin{align}
  \label{eqn:consistentILP}
  \sC = \{ M \in \pZ^{\ns \times \nt} : SM = T \}.
\end{align}
See Figure~\ref{fig:SandTandM} for the matrix formulation of $\SF$ and
$\TF$, along with one possible consistent mapping $M$ for our running example.

\subsection{Integer linear programming} \label{sec:integer_linear_program}
Finding an element of $\sC$ as defined in \refeqn{consistentILP}
corresponds to solving an integer linear program (ILP),
which is NP-hard in the worst case,
though there exist relatively effective off-the-shelf solvers such as Gurobi.
However, \emph{one} solution is not enough.
To check whether an input $x$ is in the safe set $\sF$ \refeqn{safe},
we need to check whether \emph{all} mappings $M \in \sC$ predict the same output on $x$;
that is, $x M$ is the same for all $M \in \sC$.

Our insight is that we can check whether $x \in \sF$ by solving just two ILPs.
Recall that we want to know if the output vector $x M$ can be different for different $M \in \sC$.
To do this, we pick a random vector $v \in \R^{\nt}$, and consider the scalar projection $x M v$.
The first ILP maximizes this scalar and the second one minimizes it.
If both ILPs return the same value, then with probability 1,
we can conclude that $x M$ is the same for all mappings $M \in \sC$ and thus $x \in \sF$.
The following proposition formalizes this:

\begin{proposition}
\label{prop:ilp}
Let $x$ be any input.
Let $v \sim \sN(0, I_{\nt \times \nt})$ be a random vector.
Let $a = \min_{M \in \sC} x M v$ and $b = \max_{M \in \sC} x M v$.
With probability 1, $a = b$ iff $x \in \sF$.
\end{proposition}

\begin{proof}
If $x \in \sF$, there is only one output $x M$, so $a = b$.
If $x \not\in \sF$, there exists two $M_1,M_2 \in \sC$ for which $x M_1 \neq x M_2$.
Then $w \eqdef x (M_1 - M_2) \in \R^{1 \times \nt}$ is nonzero.
The probability of $w v = 0$ is zero because the space orthogonal to $w$ is a $(\nt-1)$-dimensional space
while $v$ is drawn from a $\nt$-dimensional space.
Therefore, with probability $1$, $x M_1 v \neq x M_2 v$.
Without loss of generality, $a \le x M_1 v < x M_2 v \le b$,
so $a \neq b$.
\end{proof}

\subsection{Linear programming} 
\label{sec:linear_program}
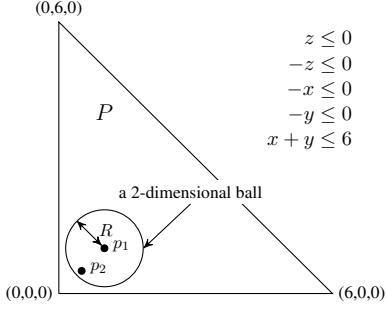
\begin{figure}
\newcommand{\scale}{0.6}
\tikzset{
	mapping/.style={
	circle,
	draw=black,
	fill=black,
	minimum size=0.15cm,
	inner sep=0,
	scale = \scale
	},
}
\centering
\begin{tikzpicture}[scale=\scale]
\draw (0,0) -- (6,0) -- (0,6) -- (0,0);
\node[anchor=west, scale=\scale] at (6,0) {(6,0,0)};
\node[anchor=south, scale=\scale] at (0,6) {(0,6,0)};
\node[anchor=east, scale=\scale] at (0,0) {(0,0,0)};

\node[mapping] (M1) at (1,1) {};
\node[anchor=west,scale=\scale] (M1name) at (M1.north east) {$p_1$};
\node[mapping] (M2) at (0.5,0.5) {};
\node[anchor=west,scale=\scale] (M2name) at (M2.north east) {$p_2$};

\node[inner sep=0,circle,draw=black,minimum size=1.7cm,scale=\scale] (p) at (1,1) {};
\draw[<->] (M1) -- (p.north west);
\node[xshift=1em,yshift=-0.3em, scale=\scale] (delta) at (p.north west) {$R$};

\node[scale=\scale] (clp) at (1,4) {\Large $P$};
\node[fill=white,yshift=1em,xshift=2em, scale=\scale] (dd) at (p.north east) {a 2-dimensional ball};
\draw[->] (dd.south) -- (p.east);

\node[scale=0.7] (equations) at (5.5,4.5) {
	\begin{tabular}{r}
	$z\le 0$\\
	$-z\le0$\\
	$-x\le0$\\
	$-y\le0$\\
	$x+y\le 6$\\
	\end{tabular}
	};
\end{tikzpicture}
\caption{
Our goal is to find two points $p_1,p_2$ in the relative interior of a polytope $P$
defined by inequalities shown on the right.
The inequalities $z \le 0$ and $-z \le 0$ are always active.
Therefore, $P$ is a $2$-dimensional polytope.
One solution to the LP \refeqn{slickLP} is
$\alpha^* =1, p^*=(1,1,0), {\xi^*}^\top=[0,0,1,1,1]$,
which results in
$p_1=(1,1,0)$ with $R = 1/\sqrt{2}$.
The other point $p_2$ is chosen randomly from the ball of radius $R$.
}
\label{fig:two_mapping_figure}
\end{figure}

\begin{algorithm*}
\caption{
  Our linear programming approach.
}
\label{alg:two_points_schema}
\newcommand{\scale}{0.9}
\scalebox{\scale}{
\begin{minipage}{9cm}
\begin{algorithmic}
\Procedure{Train}{}\\
\textbf{Input:} Training examples\\
\textbf{Output:} Generic mappings $(M_1, M_2)$
	\State Define $\sClp$ as explained in \refeqn{consistentLP}.
  \State Compute $M_1$ and a radius $R$ by solving an LP \refeqn{slickLP}.
  \State Sample $M_2$ from a ball with radius $R$ 
around $M_1$.
  \State \Return $(M_1, M_2)$
\EndProcedure
\end{algorithmic}
\end{minipage}}%
\scalebox{\scale}{
 \begin{minipage}{9cm}
\begin{algorithmic}
\Procedure{Test}{}\\
  \textbf{Input:} input $x$, mappings $(M_1, M_2)$ \\
  \textbf{Output:} A guaranteed correct $y$ or ``don't know''
  \State Compute $y_1 = xM_1$ and $y_2 = xM_2$.
  \If{$y_1=y_2$}
	 \Return $y_1$
\Else\ \Return ``don't know''
\EndIf
\EndProcedure
\end{algorithmic}
\end{minipage}}
\end{algorithm*}

\refprop{ilp} requires solving two non-trivial ILPs per input \emph{at test time}.
A natural step is to relax the integer constraint so that we solve two LPs instead. 
\begin{align}
\label{eqn:consistentLP}
\sClp &\eqdef \{ M \in \pR^{\ns \times \nt} \mid \SF M = \TF \} \\
\label{eqn:safeLP}
\sFlp &\eqdef \{ x : |\{ M(x) : M \in \sClp \}| = 1 \}.
\end{align}
The set of consistent mappings is larger ($\sClp \supseteq \sC$),
so the set of safe inputs is smaller ($\sFlp \subseteq \sF$).
Therefore, if we predict only on $\sFlp$,
we still maintain 100\% precision, although the recall could be lower.

Now we will show how to exploit the convexity of $\sClp$ (unlike $\sC$)
to avoid solving any LPs at test time at all.
The basic idea is that if we choose two mappings $M_1,M_2 \in \sClp$
``randomly enough'',
whether $xM_1 = x M_2$ is equivalent to unanimity over $\sClp$.
We could try to sample $M_1,M_2$ uniformly from $\sClp$,
but this is costly.
We instead show that ``less random'' choice suffices.
This is formalized as follows:

\begin{proposition}
	\label{prop:projection}
Let $X$ be a finite set of test inputs. 
Let $d$ be the dimension of $\sClp$.
Let $M_1$ be any mapping in $\sClp$,
and let $\text{vec}(M_2)$ be sampled from a proper density over
a $d$-dimensional ball lying in $\sClp$ centered at $\text{vec}(M_1)$.
Then, with probability 1, for all $x \in X$,
$x M_1 = x M_2$ implies $x \in \sFlp$.
\end{proposition}
\begin{proof}
We will prove the contrapositive.
If $x \not\in \sFlp$,
then $xM$ is not the same for all $M \in \sClp$.
Without loss of generality, assume not all $M \in \sClp$ agree on the $i$-th component of $xM$. 
Note that $(xM)_i=\tr(Me_i x)$, which is the inner product of $\text{vec}(M)$ and $\text{vec}(e_i x)$. 
Since $(xM)_i$ is not the same for all $M \in \sClp$ and $\sClp$ is convex,
the projection of $\sClp$ onto $\text{vec}(e_i x)$ must be a one-dimensional polytope.
For both $\text{vec}(M_1)$ and $\text{vec}(M_2)$ to have the same projection on $\text{vec}(e_i x)$,
they would have to both lie in a $(d-1)$-dimensional polytope orthogonal to $\text{vec}(e_i x)$.
Since $\text{vec}(M_2)$ is sampled from a proper density over a $d$-dimensional ball,
this has probability $0$.
\end{proof}

\newcommand{\xx}{p}
\newcommand{\Aneq}{A_{1}}
\newcommand{\bneq}{b_{1}}
\newcommand{\Aeq}{A_{0}}
\newcommand{\beq}{b_{0}}

\paragraph{Algorithm.}
We now provide an algorithm to find two points $\xx_1,\xx_2$ inside a general
$d$-dimensional polytope $P=\{\xx: A\xx \le b\}$
satisfying the conditions of \refprop{projection},
where for clarity we have simplified the notation from
$\text{vec}(M_i)$ to $\xx_i$
and $\sClp$ to $P$.

We first find a point $\xx_1$ in the relative interior of $P$,
which consists of points for which the fewest number of inequalities $j$ are active
(i.e., $a_j \xx = b_j$).
We can achieve this by solving the following LP from \citet{roundy1985identifying}:
\begin{align}
\label{eqn:slickLP}
\text{max } \bone^\top \xi \text{ s.t. } A \xx + \xi \le \alpha b, 0 \le \xi \le \bone, \alpha \ge 1.
\end{align}
Here, $\xi_j$ is a lower bound on the slack of inequality $j$,
and $\alpha$ scales up the polytope so that all the $\xi_j$ that can be positive
are exactly $1$ in the optimum solution.
Importantly, if $\xi_j = 0$, constraint $j$ is \emph{always active} for all solutions $\xx \in P$.
Let $(p^*, \xi^*, \alpha^*)$ be an optimal solution to the LP.
Then define $\Aneq$ as the submatrix of $A$ containing rows $j$ for which $\xi_j^* = 1$,
and $\Aeq$ consist of the remaining rows for which $\xi_j^* = 0$.

The above LP gives us $\xx_1 = p^* / \alpha^*$,
which lies in the relative interior of $P$ (see \reffig{two_mapping_figure}).
To obtain $\xx_2$,
define a radius $R \eqdef (\alpha \max_{j : \xi_j^* = 1} \|a_j\|_2)^{-1}$.
Let the columns of matrix $N$ form an orthonormal basis of the null space of $A_0$.
Sample $v$ from a unit $d$-dimensional ball centered at $0$, and
set $\xx_2 = \xx_1 + R N v$.

To show that $\xx_2 \in P$:
First, $\xx_2$ satisfies the always-active constraints $j$,
$a_j^\top (\xx_1 + R N v) = b_j$, by definition of null space.
For non-active $j$,
the LP ensures that $a_j^\top \xx_1 + \alpha^{-1} \le b_j$,
which implies $a_j^\top (\xx_1 + R N v) \le b_j$.

\refalg{two_points_schema} summarizes our overall procedure:
At training time, we solve a single LP \refeqn{slickLP} and draw a random vector
to obtain $M_1, M_2$ satisfying \refprop{projection}.
At test time, we simply apply $M_1$ and $M_2$,
which scales only linearly with the number of source atoms in the input.

	\begin{figure*}[t]
	\newcommand{\matindex}[1]{\mbox{\scriptsize#1}}
		\begin{center}
			\scalebox{0.65}{
				$
				\overbrace{\begin{blockarray}{cccccc}
				\matindex{\nl{area}} & \matindex{\nl{of}} & \matindex{\nl{Ohio}} & \matindex{\nl{cities}} & \matindex{\nl{in}} & \matindex{\nl{Iowa}}\\		
				\begin{block}{[cccccc]}
				1 & 1 & 0 & 0 & 0 & 1 \\
				0 & 0 & 1 & 1 & 1 & 0 \\
				0 & 0 & 0 & 1 & 1 & 1 \\
				1 & 1 & 1 & 1 & 1 & 0 \\
				\end{block}
				\end{blockarray}}^{\SF} \times M = 
				\overbrace{\begin{blockarray}{cccc}
				\matindex{\wl{area}} & \matindex{\wl{city}} & \matindex{\wl{OH}} & \matindex{\wl{IA}}\\		
				\begin{block}{[cccc]}
				1 & 0 & 0 & 1\\
				0 & 1 & 1 & 0\\
				0 & 1 & 0 & 1\\
				1 & 1 & 1 & 0\\
				\end{block}
				\end{blockarray}}^{\TF}\hspace{0.25cm}\implies\hspace{0.25cm}
				M=\overbrace{\begin{blockarray}{ccccc}
				&\matindex{\wl{area}} & \matindex{\wl{city}} & \matindex{\wl{OH}} & \matindex{\wl{IA}}\\		
				\begin{block}{c[cccc]}
				\matindex{\nl{area}}	& 1 & 0 & 0 & 0\\
				\matindex{\nl{of}} 		& 0 & 0 & 0 & 0\\
				\matindex{\nl{Ohio}}	& 0 & 0 & 1 & 0 \\
				\matindex{\nl{cities}}	& 0 & 1 & 0 & 0 \\
				\matindex{\nl{in}}		& 0 & 0 & 0 & 0\\
				\matindex{\nl{Iowa}}	& 0 & 0 & 0 & 1\\
				\end{block}
				\end{blockarray}}^{M_0}+
				\overbrace{
					\left[\begin{array}{cc}
					-1 & 0  \\           
					1  & 0  \\
					\rowcolor{yellow}
					 0  & 0  \\           
					0  & -1 \\
					0  & 1  \\  
					\rowcolor{yellow}         
					0  & 0  \\
					\end{array}\right]}^{B}
				\times
				\overbrace{\begin{bmatrix}
					a_{1,1} & a_{1,2} & a_{1,3} & a_{1,4}\\
					a_{2,1} & a_{2,2} & a_{2,3} & a_{2,4}\\
					\end{bmatrix}}^{A}		
				$	
			}
		\end{center}
		\caption{\label{fig:multiple_answer}
    Under the linear system relaxation,
    all solutions $M$ to $S M = T$ can be expressed as $M = M_0 + B A$,
    where $B$ is the basis for the null space of $S$ and $A$ is arbitrary.
    Rows $s$ of $B$ which are zero (\nl{Ohio} and \nl{Iowa})
    correspond to the safe source atoms (though not the only safe inputs).
  }
	\end{figure*} 

\subsection{Linear system} 
\label{sec:linear_system}
To obtain additional intuition about the unanimity principle,
let us relax $\sClp$ \refeqn{consistentLP} further by removing the non-negativity constraint,
which results in a linear system.
Define the relaxed set of consistent mappings to be all the solutions to the linear system and the relaxed safe set accordingly:
\begin{align}
\label{eqn:consistentLS}
\sCls &\eqdef \{ M \in \R^{\ns \times \nt} \mid \SF M = \TF \} \\
\label{eqn:safeLS}
\sFls &\eqdef \{ x : |\{ M(x) : M \in \sCls \}| = 1 \}.
\end{align}

Note that $\sCls$ is an affine subspace, so each $M \in \sCls$ can be expressed as $M_0 + B A$,
where $M_0$ is an arbitrary solution, $B$ is a basis for
the null space of $\SF$ and $A$ is an arbitrary matrix.	
Figure~\ref{fig:multiple_answer} presents the linear system for four training examples.
In the rare case that $S$ has full column rank (if we have many training examples),
then the left inverse of $S$ exists, and there is exactly one consistent mapping,
the true one ($M^* = S^\dagger T$),
but we do not require this.

Let's try to explore the linear algebraic structure in the problem.
Intuitively, if we know \nl{area of Ohio} maps to \wl{area(OH)} and
\nl{Ohio} maps to \wl{OH}, then we should conclude \nl{area of} maps to \wl{area}
by subtracting the second example from the first.
The following proposition formalizes and generalizes this intuition
by characterizing the relaxed safe set:
\begin{figure}[t]
\newcommand{\scc}{0.7}
\newcommand{\matindex}[1]{\mbox{\scriptsize#1}}
\newcommand\sdw[1]{\begin{sideways}#1\end{sideways}}

\begin{tikzpicture}[scale=\scc]
		\node[scale=\scc] (S) at (0,0) { $\overbrace{\begin{blockarray}{lccccccc}
		&\matindex{\sdw{\nl{area}}
} & \matindex{\sdw{\nl{of}}} & \matindex{\sdw{\nl{Ohio}}} & \matindex{\sdw{\nl{cities}}} & \matindex{\sdw{\nl{in}}} & \matindex{\sdw{\nl{Iowa}}}&\\	
		\begin{block}{l[cccccc]c}
		\matindex{\nl{area of Iowa}} & 1 & 1 & 0 & 0 & 0 & 1 & \color{red} +1\\
		\matindex{\nl{cities in Ohio}} & 0 & 0 & 1 & 1 & 1 & 0 &\color{red} +1\\ 
		\matindex{\nl{cities in Iowa}} & 0 & 0 & 0 & 1 & 1 & 1 &\color{red} -1\\
		\end{block}
		\cline{1-8}
		\begin{block}{l[cccccc]c}
\matindex{\nl{area of Ohio}} & 1 & 1 & 1 & 0 & 0 & 0 &\\
		\end{block}
		\end{blockarray}}^{\SF}$};
		\node[scale=\scc] (T) at (6,0) { $\overbrace{\begin{blockarray}{lccccc}
		&\matindexw{\wl{area}} & \matindexw{\wl{city}} & \matindexw{\wl{OH}} & \matindexw{\wl{IA}} &\\	
		\begin{block}{l[cccc]c}
		\matindex{\wl{area(IA)}} & 1 & 0 & 0 & 1 & \color{red} +1\\
		\matindex{\wl{city(OH)}} & 0 & 1 & 1 & 0 & \color{red} +1\\ 
		\matindex{\wl{city(IA)}} & 0 & 1 & 0 & 1 & \color{red} -1\\
		\end{block}
		\cline{1-6}
		\begin{block}{l[cccc]c}
		\matindex{\wl{area(OH)}} & 1 & 0 & 1 & 0 &\\
		\end{block}
		\end{blockarray}}^{\TF}$};
	\end{tikzpicture}
	
		\caption{\label{fig:row_space}
			Under the linear system relaxation,
			we can predict the target atoms for the new input \nl{area of Ohio}
			by adding and subtracting training examples (rows of $S$ and $T$).
		}
\end{figure}	
\begin{proposition}
	\label{prop:row_space}
	The vector $x$ is in row space of $S$ iff $x \in \sFls$.
\end{proposition}
\begin{proof}
	If $x$ is in the row space of $S$,
	we can write $x$ as a linear combination of $S$ for some coefficients $\alpha \in \R^{n}$:
	$x = \alpha^\top S$.
	Then for all $M \in \sCls$,
	we have $SM=T$, so $xM = \alpha^\top S M = \alpha^\top T$,
	which is the unique output\footnote{
    There might be more than one set of coefficients ($\alpha_1, \alpha_2$) for writing $x$. 
	However, they result to a same output: $\alpha_1^\top S=\alpha_2^\top S \implies \alpha_1^\top SM=\alpha_2^\top SM \implies \alpha_1^\top T=\alpha_2^\top T$.}
	 (See Figure~\ref{fig:row_space}).
	If $x \in \sFls$ is safe, then there exists a $y$ such that for all $M \in \sCls$, $x M = y$.
  Recall that each element of $\sCls$ can be decomposed into $M_0+BA$.
  For $x(M_0+BA)$ to be the same for each $A$,
  $x$ should be orthogonal to each column of $B$, a basis for the null space of $S$.
	This means that $x$ is in the row space of $S$.
\end{proof}
Intuitively, this proposition says that stitching new inputs together by adding and subtracting
existing training examples (rows of $S$) gives you exactly the relaxed safe set $\sFls$.

Note that relaxations increases the set of consistent mappings ($\sCls \supseteq \sClp \supseteq \sC$),
which has the contravariant effect of shrinking the safe set ($\sFls \subseteq \sFlp \subseteq \sF$).
Therefore, using the relaxation (predicting when $x \in \sFls$)
still preserves 100\% precision.

\subsection{Handling noise} 
\label{sec:remove_noise}
So far, we have assumed that our training examples are noiseless,
so that we can directly add the constraint $S M = T$.
Now assume that an adversary has made at most $\nmistakes$ additions to and
deletions of target atoms across the examples in $T$,
but of course we do not know which examples have been tainted.
Can we still guarantee 100\% precision?

The answer is yes for the ILP formulation: we simply replace the exact match condition ($S M = T$)
with a weaker one: $\|S M - T\|_1 \le \nmistakes$ (*).
The result is still an ILP, so the techniques
from Section~\ref{sec:integer_linear_program} readily apply.
Note that as $\nmistakes$ increases, the set of candidate mappings grows,
which means that the safe set shrinks.

Unfortunately, this procedure is degenerate for linear programs.
If the constraint (*) is not tight, then
$M + E$ also satisfies the constraint for any matrix $E$ of small enough norm.
This means that the consistent mappings $\sClp$ will be full-dimensional and certainly
not be unanimous on any input.

Another strategy is to remove examples from the dataset if they could be potentially noisy.
For each training example $i$, we run the ILP (*) on all but the $i$-th example.
If the $i$-th example is not in the resulting safe set~\refeqn{safe}, we remove it.
This procedure produces a noiseless dataset,
on which we can apply the noiseless linear program or linear system from the previous sections.

\section{Experiments}
\label{sec:experiments}
\subsection{Artificial data}
\newcommand{\numTrain}{120}
\newcommand{\numTest}{50}
\newcommand{\numWords}{50}
\newcommand{\numClusters}{10} 
\newcommand{\numPredicates}{20}
\newcommand{\minSentenceLength}{5}
\newcommand{\maxSentenceLength}{10}
\newcommand{\minAWordCanGo}{0}
\newcommand{\maxAWordCanGo}{2}

\newcommand{\pOne}{0.2}
\newcommand{\pTwo}{0.5}
\newcommand{\pThree}{0.7}
\newcommand{\nTrial}{100}

\begin{figure*}
\newcommand{\scale}{0.6}
\newcommand{\szn}{0.22}
   \pgfplotsset{
    small,
    legend style={
        at={(0.01,0.01)},
        anchor=south west,
		scale = 0.25,
		font = \tiny,
    },
   }%
\begin{subfigure}[t]{\szn\textwidth}
\centering
\begin{tikzpicture}[baseline,scale=\scale]
   \begin{axis}[
   xmax=1.1,xmin=0,
   ymin= 0,ymax=1.1,
   xlabel=\emph{Fraction of training data},
ylabel=\emph{Recall},
    y label style={at={(axis description cs:0.15,0.5)},anchor=south},
   xtick={0,0.2,0.4,...,1},
   ytick={0,0.2,0.4,...,1},
    legend pos= south east,
 ymajorgrids=true,
    grid style=dashed,
legend cell align=left,
   ]
\addplot coordinates{
(0.1,1.0)
(0.2,1.0)
(0.30000000000000004,1.0)
(0.4,1.0)
(0.5,1.0)
(0.6,1.0)
(0.7,1.0)
(0.7999999999999999,1.0)
(0.8999999999999999,1.0)
(0.9999999999999999,1.0)
};
\addplot coordinates{
(0.1,0.0)
(0.2,0.08)
(0.30000000000000004,0.5)
(0.4,0.56)
(0.5,0.56)
(0.6,0.56)
(0.7,0.62)
(0.7999999999999999,0.94)
(0.8999999999999999,0.94)
(0.9999999999999999,1.0)
};
\addplot coordinates{
(0.1,0.0)
(0.2,0.44)
(0.30000000000000004,0.56)
(0.4,0.62)
(0.5,0.68)
(0.6,1.0)
(0.7,1.0)
(0.7999999999999999,1.0)
(0.8999999999999999,1.0)
(0.9999999999999999,1.0)
};
\addplot coordinates{
(0.1,0.04)
(0.2,0.46)
(0.30000000000000004,0.58)
(0.4,0.66)
(0.5,0.72)
(0.6,1.0)
(0.7,1.0)
(0.7999999999999999,1.0)
(0.8999999999999999,1.0)
(0.9999999999999999,1.0)
};

    \legend{precision (all), recall (ILP), recall (LP), recall (LS)}
00    \end{axis}
    \end{tikzpicture}
\caption{\label{fig:artificial_all_models_experiment} All the relaxations reach
  100\% recall; relaxation results in slightly slower convergence.}
\end{subfigure}\quad%
\begin{subfigure}[t]{\szn\textwidth}
\begin{tikzpicture}[baseline,scale=\scale]
   \begin{axis}[
   xmax=152,xmin=0,
   ymin= 0,ymax=1.1,
   xlabel=$\nmistakes$,
    y label style={at={(axis description cs:0.15,0.5)},anchor=south},
   xtick={0,30,60,...,150},
   ytick={0,0.2,0.4,...,1},
    legend pos= south west,
 ymajorgrids=true,
    grid style=dashed,
legend cell align=left,
   ]
\addplot coordinates{
	(0,    1.0)
	(30,    1.0)
	(60,    1.0)
	(90,    1.0)
	(120,    1.0)
	(150,    1.0)

};

\addplot coordinates{
	(0,    1.0)
	(30,    1.0)
	(60,    0.84)
	(90,    0.28)
	(120,    0.28)
	(150,    0.0)
};
    \legend{precision (ILP), recall (ILP)}
    \end{axis}
    \end{tikzpicture}
\caption{\label{fig:artificial_noise_experiment} Size of the safe set shrinks with
  increasing number of mistakes in the training data.}
\end{subfigure}\quad%
\begin{subfigure}[t]{0.45\textwidth}
   \begin{tikzpicture}[baseline,scale=\scale]
   \begin{axis}[
   xmax=1.1,xmin=0,
   ymin= 0.3,ymax=1.1,
   xlabel=\emph{Recall},
ylabel=\emph{Precision},
    y label style={at={(axis description cs:0.15,0.5)},anchor=south},
   xtick={0,0.2,0.4,...,1},
   ytick={0,0.2,0.4,...,1},
legend style={at={(1.0,0.0)},anchor=south east},
    legend columns=2, 
 ymajorgrids=true,
legend cell align=right,
    grid style=dashed,   
   ]
\addplot coordinates{
(0.04,1.0)
};
\addplot coordinates{
(0.04,0.5)
(0.1,0.5555555555555556)
(0.24,0.5714285714285714)
(0.26,0.5)
(0.28,0.4375)
(0.3,0.3488372093023256)
};
\addplot coordinates{
(0.48,1.0)
};
\addplot coordinates{
(0.48,0.8571428571428571)
(0.48,0.8571428571428571)
(0.5,0.8333333333333334)
(0.5,0.8064516129032258)
(0.56,0.717948717948718)
(0.64,0.64)
};
\addplot coordinates{
(0.78,1.0)
};
\addplot coordinates{
(0.78,0.9512195121951219)
(0.78,0.9512195121951219)
(0.78,0.9512195121951219)
(0.78,0.9069767441860465)
(0.82,0.8913043478260869)
(0.84,0.84)
};
  \legend{
	$\pointM$(\pOne), 
 	$\ourM$\ (\pOne), 
 	$\pointM$(\pTwo), 
 	$\ourM$\ (\pTwo),
 	$\pointM$(\pThree), 
 	$\ourM$\ (\pThree)}
    \end{axis}
\end{tikzpicture}%
   \begin{tikzpicture}[baseline,scale=\scale]
   \begin{axis}[
   xmax=1.1,xmin=0,
   ymin= 0.3,ymax=1.1,
   xlabel=\emph{Recall},
ylabel=\emph{Precision},
    y label style={at={(axis description cs:0.15,0.5)},anchor=south},
   xtick={0,0.2,0.4,...,1},
   ytick={0,0.2,0.4,...,1},
legend style={at={(1.0,0.0)},anchor=south east},
    legend columns=2, 
 ymajorgrids=true,
legend cell align=left,
    grid style=dashed,   
   ]

\addplot coordinates{
(0.04,1.0)
};
\addplot coordinates{
(0.2,1.0)
(0.2,1.0)
(0.2,0.9090909090909091)
(0.22,0.7857142857142857)
(0.24,0.5454545454545454)
(0.24,0.35294117647058826)
};
\addplot coordinates{
(0.52,1.0)
};
\addplot coordinates{
(0.62,1.0)
(0.62,1.0)
(0.7,1.0)
(0.7,0.9459459459459459)
(0.7,0.8974358974358975)
(0.72,0.8372093023255814)
};
\addplot coordinates{
(0.78,1.0)
};
\addplot coordinates{
(0.8,1.0)
(0.8,1.0)
(0.8,1.0)
(0.82,1.0)
(0.82,0.9111111111111111)
(0.84,0.84)
};

 \legend{
	$\pointM$(\pOne), 
 	$\ourM$\ (\pOne), 
 	$\pointM$(\pTwo), 
 	$\ourM$\ (\pTwo),
 	$\pointM$(\pThree), 
 	$\ourM$\ (\pThree)}

  \end{axis}
\end{tikzpicture}
\caption{\label{fig:artificial_adversary_experiment}
  Performance of the point estimate ($\pointM$) and unanimous prediction ($\ourM$)
when the inputs are chosen adversarially for $\pointM$ (left) and for $\ourM$ (right).}
\end{subfigure}
\caption{\label{fig:artificial_experiments} Our algorithm always obtains
100\% precision with
(a) different amounts of training examples and different relaxations,
(b) existence of noise, and
(c) adversarial input distributions.
}
\end{figure*}
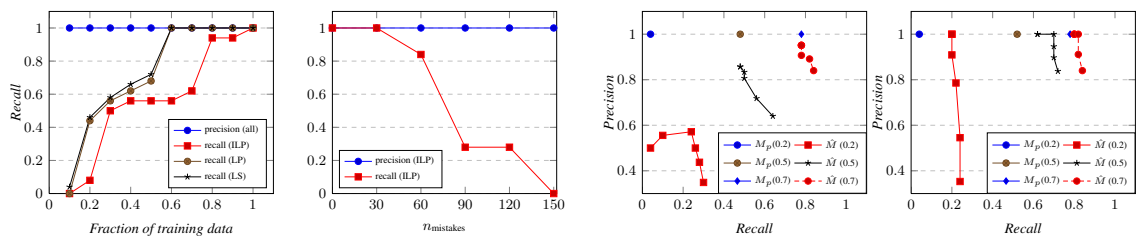

We generated a true mapping $M^*$
from \numWords{} source atoms to \numPredicates{} target atoms
so that each source atom maps to \minAWordCanGo{}--\maxAWordCanGo{} target atoms.
We then created \numTrain{} training examples and \numTest{} test examples,
where the length of every input is between \minSentenceLength{} and \maxSentenceLength{}. 
The source atoms are divided into \numClusters{} clusters,
and each input only contains source atoms from one cluster.

\reffig{artificial_all_models_experiment} shows the results for
$\sF$ (integer linear programming),
$\sFlp$ (linear programming), and
$\sFls$ (linear system).
All methods attain 100\% precision, and as expected, relaxations lead to lower recall,
though they all can reach 100\% recall given enough data.

\paragraph{Comparison with point estimation.}
Recall that the unanimity principle $\ourM$ reasons over the entire set of consistent mappings,
which allows us to be robust to changes in the input distribution,
e.g., from training set attacks \cite{mei2015teaching}.
As an alternative, consider computing the point estimate $\pointM$ that minimizes $\|SM-T\|_2^2$
(the solution is given by $\pointM = S^\dagger T$).
The point estimate, by minimizing the average loss,
implicitly assumes i.i.d.~examples.
To generate output for input $x$ we compute $y=x\pointM$ and round each
coordinate $y_t$ to the closest integer.
To obtain a precision-recall tradeoff,
we set a threshold $\epsilon$
and if for all target atoms $t$, the interval $[y_t-\epsilon, y_t+\epsilon)$ contains an integer,
we set $y_t$ to that integer; otherwise we report ``don't know'' for input $x$.

To compare unanimous prediction $\ourM$ and point estimation $\pointM$,
for each $f \in \{ \pOne{}, \pTwo{}, \pThree \}$,
we randomly generate \nTrial{} subsampled datasets consisting of an $f$ fraction of the training examples.
For $\pointM$, we sweep $\epsilon$ across $\{0.0, 0.1, \dots, 0.5\}$ to obtain a ROC curve. 
In \reffig{artificial_adversary_experiment}(left/right),
we select the distribution that results in the maximum/minimum
difference between $F_1(\ourM)$ and $F_1(\pointM)$ respectively.
As shown, $\ourM$ has always 100\% precision,
while $\pointM$ can obtain less 100\% precision over its full ROC curve.
An adversary can only hurt the recall of unanimous prediction.

\paragraph{Noise.}
As stated in \refsec{remove_noise}, our algorithm has the ability to guarantee
100\% precision even when the adversary can modify the outputs.
As we increase the number of predicate additions/deletions ($\nmistakes$),
\reffig{artificial_noise_experiment} shows that
precision remains at 100\%, while recall naturally decreases in response to being less
confident about the training outputs.

\subsection{Semantic parsing on GeoQuery}
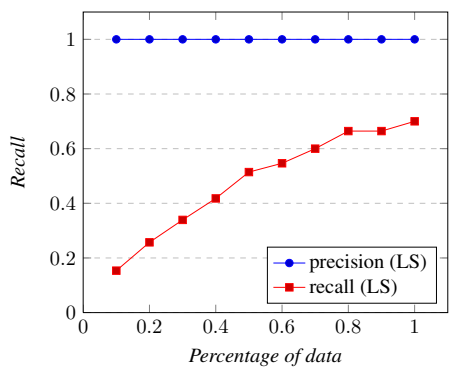
\begin{figure}
	\newcommand{\scale}{0.65} 

			\begin{tikzpicture}[baseline,scale=0.7]
   \begin{axis}[
   xmax=1.1,xmin=0,
   ymin= 0,ymax=1.1,
   xlabel=\emph{Percentage of data},
ylabel=\emph{Recall},
   xtick={0,0.2,0.4,...,1},
   ytick={0,0.2,0.4,...,1},
    legend pos= south east,
 ymajorgrids=true,
    grid style=dashed,
legend cell align=left,
   ]
\addplot coordinates{
(0.1,1.0)
(0.2,1.0)
(0.30000000000000004,1.0)
(0.4,1.0)
(0.5,1.0)
(0.6,1.0)
(0.7,1.0)
(0.7999999999999999,1.0)
(0.8999999999999999,1.0)
(0.9999999999999999,1.0)
};

\addplot coordinates{
(0.1,0.15357142857142858)
(0.2,0.2571428571428571)
(0.30000000000000004,0.3392857142857143)
(0.4,0.41785714285714287)
(0.5,0.5142857142857142)
(0.6,0.5464285714285714)
(0.7,0.6)
(0.7999999999999999,0.6642857142857143)
(0.8999999999999999,0.6642857142857143)
(0.9999999999999999,0.7)
};
    \legend{precision (LS), recall (LS)}    \end{axis}
    \end{tikzpicture}
			\caption{We maintain 100\% precision while recall increases with the number of training examples.}
			\label{fig:main_result_geo}
\end{figure}

\label{sec:geo_query_experiments}

We now evaluate our approach on the standard GeoQuery dataset \cite{zelle96geoquery},
which contains 880 utterances and their corresponding logical forms. 
The utterances are questions related to the US geography,
such as: \nlq{what river runs through the most states}.

We use the standard 600/280 train/test split \citep{zettlemoyer05ccg}. 
After replacing entity names by their types\footnote{If an entity name has more
than one type we replace it by concatenating all of its possible types.} based
on the standard entity lexicon, there are 172 different words and 57 different
predicates in this dataset.

\paragraph{Handling context.}
Some words are polysemous in that they map to two predicates:
in \nlq{largest river} and \nlq{largest city},
the word \nl{largest} maps to \wl{longest} and \wl{biggest}, respectively. 
Therefore, instead of using words as source atoms,
we use bigrams,
so that each source atom always maps to the same target atoms.

\begin{table*}[ht]
\begin{center}
\newcommand{\scale}{0.64}
\newcommand{\x}{15}
\newcommand{\y}{3}
\centering
\begin{tikzpicture}[scale=\scale]

\node[scale=\scale]  (utterances) at (0,0) 
{\begin{tabular}{|c|c|c|c|c|}
\hline
\bf utterances &\bf logical form (A) &\bf target atoms (A) &\bf logical form (B) & \bf target atoms (B) \\ \hline

cities traversed by the Columbia & \wl{city(x),loc(x,Columbia)} &\wl{city,loc,Columbia}
& \wl{city(loc\_1(Columbia))} & \wl{city,loc\_1,Columbia} \\ \hline

cities of Texas & \wl{city(x),loc(Texas,x)} & 
\wl{city,loc,Texas} & \wl{city(loc\_2(Texas))} & 
\wl{city,loc\_2,Texas} \\ \hline
\end{tabular}};

\end{tikzpicture}
\end{center}
\caption{\label{tab:compositional_target_atoms}
  Two different choices of target atoms:
  (A) shows predicates and (B) shows predicates conjoined with their argument position.
  (A) is sufficient for simply recovering the predicates,
  whereas (B) allows for logical form reconstruction.
}
\end{table*}

\paragraph{Reconstructing the logical form.}
We define target atoms to include more information than just the predicates,
which enables us to reconstruct logical forms from the predicates.
We use the variable-free functional logical forms \cite{kate05funql},
in which each target atom is a predicate conjoined with its argument order
(e.g., \wl{loc\_1} or \wl{loc\_2}).
\reftab{compositional_target_atoms} shows two different choices of target atoms.
At test time, we search over all possible ``compatible'' ways
of combining target atoms into logical forms.
If there is exactly one, then we return that logical form and abstain otherwise.
We call a predicate combination ``compatible'' if it appears in the training set.

We put a ``null'' word at the end of each sentence,
and collapsed the \wl{loc} and \wl{traverse} predicates.
To deal with noise, we minimized $\|SM-T\|_1$ over real-valued mappings and
removed any example (row) with non-zero residual.
We perform all experiments using the linear system relaxation.
Training takes under 30 seconds.

\reffig{main_result_geo} shows precision and recall
as a function of the number of the training examples. 
We obtain 70\% recall over predicates on the test examples.
84\% of these have a unique compatible way of combining target atoms into a logical form,
which results in a 59\% recall on logical forms.

Though our modeling assumptions are incorrect for real data, we were still able to get 100\%
precision for all training examples.
Interestingly, the linear system (which allows negative mappings) helps model GeoQuery dataset better
than the linear program (which has a non-negativity constraint).
There exists a predicate \wl{all:e} in GeoQuery
that is in every sentence unless the utterance contains a proper noun. 
With negative mappings,
\nl{null} maps to \wl{all:e},
while each proper noun maps to its proper predicate \emph{minus} \wl{all:e}. 

There is a lot of work in semantic parsing that tackles the GeoQuery dataset
\citep{zelle96geoquery,zettlemoyer05ccg,wong07synchronous,kwiatkowski10ccg,liang11dcs},
and the state-of-the-art is 91.1\% precision and recall \citep{liang11dcs}.
However, none of these methods can guarantee 100\% precision, and they perform more
feature engineering, so these numbers are not quite comparable.
In practice, one could use our unanimous prediction approach in conjunction with others:
For example, one could run a classic semantic parser and simply certify 59\% of the examples
to be correct with our approach.
In critical applications, one could use our approach as a first-pass filter,
and fall back to humans for the abstentions.

\section{Extensions}
\label{sec:other_applications}
\subsection{Learning from denotations}
\label{sec:learning_from_denotations}
Up until now, we have assumed that we have input-output pairs.
For semantic parsing, this means annotating sentences with logical forms (e.g., \nl{area of Ohio} to \wl{area(OH)})
which is very expensive.
This has motivated previous work to learn from question-answer pairs
(e.g., \nl{area of Ohio} to \wl{44825}) \citep{liang11dcs}.
This provides weaker supervision:
For example, \wl{44825} is the area of Ohio (in squared miles), but
it is also the zip code of Chatfield.
So, the true output could be either \wl{area(OH)} or \wl{zipcode(Chatfield)}.

In this section, we show how to handle this form of weak supervision by asking
for unanimity over additional selection variables.
Formally, we have $D=\{(x_1,Y_1), \dots, (x_n, Y_n)\}$ as a set of training examples,
here each $Y_i$ consists of $k_i$ candidate outputs for $x_i$.
In this case, the unknowns are the mapping $M$ as before along with a selection vector $\pi_i$,
which specifies which of the $k_i$ outputs in $Y_i$ is equal to $x_i M$.
To implement the unanimity principle, we need to consider the set of all
consistent solutions $(M, \pi)$.

We construct an integer linear program as follows:
Each training example adds a constraint that the
output of it should be exactly one of its candidate output. 
For the $i$-th example, we form a matrix $T_i \in \R^{k_i \times \nt}$
with all the $k_i$ candidate outputs.
Formally we want $x_i M = \pi_i T_i$.  The entire ILP is:
\begin{center}
\begin{tabular}{l}
$\forall i,\ x_i M = \pi_i T_i$ \\ 
$\forall i,\ \sum_j \pi_{ij} = 1$ \\
$\pi, M \ge 0$  \\
\end{tabular}
\end{center}
Given a new input $x$,
we return the same output if $xM$ is same for all consistent solutions $(M, \pi)$.
Note that we can effectively ``marginalize out'' $\pi$.
We can also relax this ILP into an linear program following Section~\ref{sec:linear_program}.

\subsection{Active learning}
\begin{figure}
				\begin{tikzpicture}[baseline,scale=0.65]
   \begin{axis}[
   xmax=1.1,xmin=0,
   ymin= 0,ymax=1.1,
   xlabel=\emph{Percentage of data},
ylabel=\emph{Recall},
   xtick={0,0.2,0.4,...,1},
   ytick={0,0.2,0.4,...,1},
    legend pos= south east,
 ymajorgrids=true,
    grid style=dashed,
legend cell align=left,
   ]
\addplot coordinates{
(0.1,0.18214285714285713)
(0.2,0.32142857142857145)
(0.30000000000000004,0.46785714285714286)
(0.4,0.6)
(0.5,0.6857142857142857)
(0.6,0.7)
(0.7,0.7)
(0.7999999999999999,0.7)
(0.8999999999999999,0.7)
(0.9999999999999999,0.7)
};

\addplot coordinates{
(0.1,0.15357142857142858)
(0.2,0.2571428571428571)
(0.30000000000000004,0.3392857142857143)
(0.4,0.41785714285714287)
(0.5,0.5142857142857142)
(0.6,0.5464285714285714)
(0.7,0.6)
(0.7999999999999999,0.6642857142857143)
(0.8999999999999999,0.6642857142857143)
(0.9999999999999999,0.7)
};

    \legend{Active learning, Passive learning}
  \end{axis}
    \end{tikzpicture}
				\caption{When we choose examples to be linearly independent,
					we only need half the number of examples 
					to achieve the same performance.}
				\label{fig:active_learning}
\end{figure}
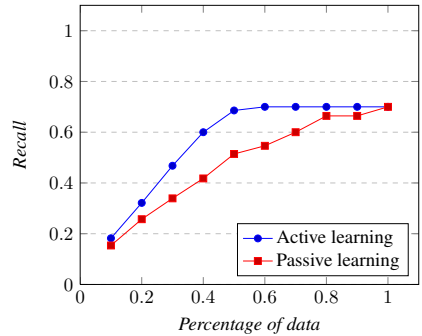
\label{sec:active_learning}
A side benefit of the linear system relaxation (Section~\ref{sec:linear_system})
is that it suggests an active learning procedure.
The setting is that we are given a set of inputs (the matrix $S$),
and we want to (adaptively) choose which inputs (rows of $S$) to obtain
the output (corresponding row of $T$) for.

Proposition~\ref{prop:row_space} states that under the linear system formulation,
the set of safe inputs $\sFls$ is exactly the same as the row space of $S$.
Therefore, if we ask for an input that is already in the row space of $S$,
this will not affect $\sFls$ at all.
The algorithm is then simple:
go through our training inputs $x_1, \dots, x_n$ one by one
and ask for the output only if it is not in the row space of the previously-added inputs $x_1, \dots, x_{i-1}$.

\reffig{active_learning} shows the recall
when we choose examples to be linearly independent in this way in comparison to when we choose
examples randomly.
The active learning scheme requires half as many labeled examples as the
passive scheme to reach the same recall.
In general, it takes $\text{rank}(S) \le n$ examples
to obtain the same recall as having labeled all $n$ examples.
Of course, the precision of both systems is 100\%.

\subsection{Paraphrasing}
\label{sec:paraphrasing}

Another side benefit of the linear system relaxation (\refsec{linear_system})
is that we can easily partition the safe set $\sFls$ \refeqn{safeLS} into subsets of
utterances which are paraphrases of each other.
Two utterances are paraphrase of each other if both map to the same logical form,
e.g., \nlq{Texas's capital} and \nlq{capital of Texas}.
Given a sentence $x \in \sFls$, our goal is to find all of its paraphrases in $\sFls$. 

As explained in \refsec{linear_system}, we can represent each input
$x$ as a linear combination of $S$ for some coefficients $\alpha \in \R^{n}$:
$x = \alpha^\top S$.
We want to find all $x' \in \sFls$ such that $x'$ is guaranteed to map to
the same output as $x$. 
We can represent $x'=\beta^\top S$ for some coefficients $\beta \in \R^{n}$.
The outputs for $x$ and $x'$ are thus $\alpha^\top T$ and $\beta^\top T$, respectively. 
Thus we are interested in
$\beta$'s such that $\alpha^\top T = \beta^\top T$,
or in other words, $\alpha - \beta$ is in the null space of $T^\top$.
Let $B$ be a basis for the null space of $T^\top$.
We can then write $\alpha - \beta = B v$ for some $v$.
Therefore, the set of paraphrases of $x\in \sFls$ are:
\begin{align}
\label{eqn:paraphrase}
\text{Paraphrases}(x) \eqdef \{ (\alpha - B v)^\top S : v \in \R^n \}.
\end{align}

\section{Discussion and related work}
\label{sec:conclusion}
Our work is motivated by the semantic parsing task
(though it can be applied to any set-to-set prediction task).
Over the last decade, there has been much work on semantic parsing,
mostly focusing on learning from weaker supervision \citep{liang11dcs,goldwasser11confidence,artzi11conversations,artzi2013weakly},
scaling up beyond small databases \citep{cai2013large,berant2013freebase,pasupat2015compositional},
and applying semantic parsing to other tasks \citep{matuszek2012grounded,kushman2013regex,artzi2013weakly}.
However, only \citet{popescu03precise} focuses on precision.
They also obtain 100\% precision,
but with a hand-crafted system,
whereas we \emph{learn} a semantic mapping.

The idea of computing consistent hypotheses appears in the classic theory of version spaces
for binary classification \citep{mitchell1977version} and has been extended to
more structured settings \citep{vanlehn1987version,lau2000version}.
Our version space is used in the context of the unanimity principle,
and we explore a novel linear algebraic structure.
Our ``safe set'' of inputs appears in the literature as the complement of the
disagreement region \citep{hanneke2007bound}.  
They use this notion for active learning,
whereas we use it to support unanimous prediction.

There is classic work on learning classifiers that can abstain
\citep{chow1970optimum,tortorella2000optimal,balsubramani2016learning}.
This work, however, focuses on the classification setting,
whereas we considered more structured output settings (e.g., for semantic parsing).
Another difference is that we operate in a more adversarial setting
by leaning on the unanimity principle.

Another avenue for providing user confidence is probabilistic calibration
\citep{platt1999probabilistic}, which has been explored more recently for
structured prediction \citep{kuleshov2015calibrated}.  However, these methods
do not guarantee precision for \emph{any} training set and test input.

In summary, we have presented the unanimity principle for guaranteeing 100\% precision.
For the task of learning semantic mappings,
we leveraged the linear algebraic structure in our problem to
make unanimous prediction efficient.
We view our work as a first step in learning reliable semantic parsers.
A natural next step is to explore our framework
with additional modeling improvements---especially in dealing with
context, structure, and noise.

\paragraph{Reproducibility.} All code, data, and experiments for this paper are available on the CodaLab platform at
{\small \sloppy \url{https://worksheets.codalab.org/worksheets/
0x593676a278fc4e5abe2d8bac1e3df486/}}.

\paragraph{Acknowledgments.}
We would like to thank the anonymous reviewers for their helpful comments.
We are also grateful for
a Future Of Life Research Award and NSF grant CCF-1138967,
which supported this work.

\bibliographystyle{acl2016}
\bibliography{main}

\end{document}